\documentclass[runningheads]{llncs}
\usepackage{graphicx}
\usepackage{marvosym}
\usepackage{multirow}
\usepackage{booktabs}
\usepackage{hyperref}
\usepackage{amsmath, amssymb}
\usepackage[table,xcdraw]{xcolor}
\newcolumntype{P}[1]{>{\centering\arraybackslash}p{#1}}

\begin{document}
\title{DRILL: Dynamic Representations for Imbalanced Lifelong Learning}
%
%
\author{Kyra Ahrens (\Letter) \and
Fares Abawi \and
Stefan Wermter}


%
\authorrunning{K. Ahrens et al.}
%
\institute{Knowledge Technology, Department of Informatics, University of Hamburg
\email{\{kyra.ahrens,fares.abawi,stefan.wermter\}@uni-hamburg.de}\\
\url{www.knowledge-technology.info}}
\maketitle              
\begin{abstract}
Continual or lifelong learning has been a long-standing challenge in machine learning to date, especially in natural language processing (NLP). Although state-of-the-art language models such as BERT have ushered in a new era in this field due to their outstanding performance in multitask learning scenarios, they suffer from forgetting when being exposed to a continuous stream of non-stationary data. In this paper, we introduce DRILL, a novel lifelong learning architecture for open-domain sequence classification. DRILL leverages a biologically inspired self-organizing neural architecture to selectively gate latent language representations from BERT in a domain-incremental fashion. We demonstrate in our experiments that DRILL outperforms current methods in a realistic scenario of imbalanced classification from a data stream without prior knowledge about task or dataset boundaries. To the best of our knowledge, DRILL is the first of its kind to use a self-organizing neural architecture for open-domain lifelong learning in NLP.

\end{abstract}
%
%
%

\section{Introduction}

Humans possess the ability to continuously acquire, reorganize, integrate, and enrich linguistic concepts throughout their lives. As early as infancy and based on an innate inference capability, the conventional symbols of language are learned within a socio-communicative context. The underlying neuro-cognitive mechanisms involved in human language acquisition are still far from being fully understood. However, they offer great potential for computational models that are inspired by the neuroanatomical mechanisms in the mammalian brain, enabling the continual integration of consolidated linguistic knowledge with current experience~\cite{tomasello1992social}. 

With the advent of deep learning and the surge of computational resources and data collection, state-of-the-art transformer-based language models (LM) such as BERT~\cite{Devlin2019} and OpenAI GPT~\cite{Radford2018} have gradually moved away from the symbolic level and given way to \textit{isolated learning} solutions revealing an outstanding performance on downstream NLP tasks in a multitask set-up~\cite{chen2018lifelong}. Yet when being exposed to a sequence of tasks,  \textit{catastrophic forgetting} or \textit{catastrophic interference} of previously learned concepts was observed~\cite{MCCLOSKEY1989109}. As re-training on all prior data would be inefficient both in terms of computational cost and memory capacity, this observation motivated the introduction of \textit{continual} or \textit{lifelong language learning} (LLL).

Despite recent advances in LLL, most current methods make overly simplistic assumptions that are in stark contrast to realistic, biologically inspired learning settings. This includes enabling multiple passes over the input data stream instead of single-epoch training, resulting in a surge of computational cost. Such methods further rely on perfectly balanced and annotated data, arranged in a way that the assumption of independent and identically distributed samples holds. As a consequence, they are poorly applicable to few-shot, unsupervised, or self-supervised learning scenarios~\cite{Biesialska2020}. 

Thus, striving for a more biologically grounded model architecture and training set-up, we introduce DRILL, a text classification model applicable to CL settings that involve the presence of a continuous stream of imbalanced data without prior knowledge about task boundaries or probability distributions. DRILL is a hybrid architectural and rehearsal-based CL method that uses meta-learning and a self-organizing neural architecture to enable rapid adaptation to novel data while minimizing catastrophic forgetting.

Due to the lack of a continual text classification benchmark of imbalanced data, we introduce two sampling strategies to induce class imbalance artificially. These strategies are evaluated on five text classification datasets presented by Zhang et al.~\cite{zhang2015}, commonly used as CL benchmarks in NLP. With this setting, we show in our experiments that our model outperforms current baselines while better generalizing to unseen data.

\section{Related Work}

\subsection{Continual Learning}

Striving for a balance between memory consolidation and generalization to new input data from non-stationary distributions, also referred to as the \textit{stability-plasticity dilemma}~\cite{Grossberg1980}, paved the way for various LLL approaches in recent years. These approaches can be fully or partially categorized into regularization, rehearsal, and dynamic architectures: 

\textit{Regularization-based} approaches constrain the plasticity of a learning model either by introducing additional loss terms for weight adaptation at a fixed model capacity~\cite{Kirkpatrick2017,Zenke2017}, or by setting an additional constraint on prior tasks' predictions to be kept invariant using \textit{knowledge distillation}~\cite{Li2018LwF}. This fixed-capacity paradigm contrasts with \textit{architecture-based} CL models that assign some model capacity to each task and therefore dynamically expand in response to novel input~\cite{Parisi2019,rusu2016progressive}. Inspired by the concept of memory consolidation, \textit{rehearsal-based} (or \textit{memory replay}) approaches maintain performance on prior tasks by storing and retraining the model on old training samples from an episodic memory~\cite{chaudhry2019agem,dautume2019mbpa++,Rebuffi2017}. 

To limit the associated memory overhead with an increasing number of tasks, \textit{pseudo-rehearsal} approaches employing generative network architectures have been proposed. Such models rely on experience replay of task-representative samples or latent representations based on statistical properties learned from old training data~\cite{kemker2018fearnet,Parisi2018}. Two such generative replay approaches based on GPT-2~\cite{Radford2018}, i.e. Language Modelling for Lifelong Language Learning (LAMOL)~\cite{sun2019lamol} and Distill and Replay (DnR)~\cite{sun2020distill} view LLL through the lens of question answering. DnR deviates from LAMOL in that it bounds model complexity through knowledge distillation following a teacher-student strategy. Although both approaches are the current performance leaders on datasets benchmarked in this work, they require multiple epochs of training and explicit knowledge about task boundaries. Such preconditions deviate from the realistic CL scenario we advocate for in this work.

\subsection{Meta-Learning}
\label{meta-learning}

Meta-learning~\cite{thrun1998meta} has become increasingly popular in recent years as it paved the way for sophisticated algorithms capable of quickly adapting to new data. Online aware Meta-Learning (OML)~\cite{Javed2019} combines the common meta-learning objective of maximizing fast adaptation to new tasks with the CL objective of minimizing catastrophic interference during training. A Neuromodulated Meta-Learning Algorithm (ANML)~\cite{Beaulieu2020} extends OML by an independent representation learning stream to selectively gate latent activations. Holla et al.~\cite{holla2020} introduce a sparse experience replay mechanism to OML and ANML, denoting their two novel methods by OML-ER and ANML-ER respectively. Both extensions outperform state-of-the-art methods for text classification and question answering benchmarks under a training set-up in which data becomes only available over time and a lack of information about when a dataset or task boundary is crossed.  

\subsection{Growing Memory and Self-Organization}

In an attempt to mimic the explicit memory formation in the mammalian brain, early artificial neural networks based on competitive learning mechanisms and self-organization have been developed and refined~\cite{fritzke1995gng,kohonen1990som}. One more recent extension to such topology learning methods is the Self-organizing Incremental Neural Network (SOINN) algorithm, which regulates plasticity in unsupervised learning tasks by means of dynamically creating, adapting, and deleting neurons~\cite{shen2010soinn}. SOINN+~\cite{Wiwatcharakoses2020} extends the original SOINN algorithm by introducing a novel node deletion mechanism based on (i) idle time, (ii) trustworthiness, and (iii) non-usage of a network unit. Given that SOINN+ successfully demonstrates its resilience to noisy data and its ability to learn a high-quality topology from the input domain while keeping the number of nodes small, we utilize it as a semantic memory component in our DRILL architecture. 

\section{Methods}
\label{methods}

With the challenge of achieving LLL from unbalanced data in mind, we lay the theoretical foundation for our proposed DRILL method.

\subsection{Task Formulation}

Consider an ordered sequence of tasks $\mathcal{T}=\{T_1, T_2, \dots, T_N\}$, where we observe $n_k$ annotated input samples from the $k$-th task, i.e. $T_k = \{(\boldsymbol{x}^i_{k}, y^i_{k})\}^{n_k}_{i=1}$ drawn from the distribution $P_k(\mathcal{X}, \mathcal{Y})$. Assuming a realistic scenario of missing task and dataset descriptors, we have no knowledge about which task each input sample belongs to. Following prior work~\cite{holla2020}, we define task in terms of text classification domain, i.e. sentiment, news topic, question-and-answer, and ontology. Our objective is to learn a model $f_{\theta}: \mathcal{X} \to \mathcal{Y}$ with parameters $\theta$ to minimize the negative log-likelihood averaged across all $N$ tasks
\begin{equation}
    \mathcal{L}(\theta) = - \frac{1}{N} \sum_{k=1}^{N} \text{ln} \, P(\boldsymbol{x}_k \,| \, y_k \, ; \, \theta ) 
\end{equation}

\subsection{Progressive Imbalancing}
\label{progressiveimbalancing}

Prior work on lifelong text classification \cite{holla2020,dautume2019mbpa++,sun2019lamol,sun2020distill} has traditionally deployed the perfectly balanced version of the five NLP datasets by Zhang et al.~\cite{zhang2015}. Following the idea of d'Autume et al.~\cite{dautume2019mbpa++}, we introduce two sampling techniques called \textit{progressive reduction} ($R$) and \textit{progressive expansion} ($E$), which exponentially increase or decrease the number of samples for each incoming task, such that 
\begin{equation}
    n^{R}_{k+1} \leftarrow \left\lfloor \frac{n^{R}_{k}}{2} \right\rfloor
\end{equation}
with progressive reduction and
\begin{equation}
    n^{E}_{k+1} \leftarrow 2 \cdot n^{E}_k
\end{equation}
with progressive expansion respectively, and $k \in \{1, \dots, N\}$. Both sampling techniques allow us to simulate two opposite LLL settings in which data at an early or late stage are significantly less present.  

\subsection{Episode Generation}

For the construction of training episodes and experience rehearsal from episodic memory, we follow a commonly adopted set-up~\cite{holla2020,dautume2019mbpa++}:

Under the assumption that samples arrive in batches of size $s$ and are written into episodic memory module $\mathcal{M}_{\mathcal{E}}$ with probability $p_{\mathcal{E}}$, we construct the $i$-th episode from $b$ batches, where the first $b-1$ batches denote support set $\mathcal{S}_i$ and the $b$-th batch denotes query set $\mathcal{Q}_i$. 

After having observed $R_{I}$ samples from the stream, $\left\lfloor r \cdot R_I \right\rfloor$ samples from $\mathcal{M}_{\mathcal{E}}$ are randomly being drawn for rehearsal, where $r \in [0, 1]$ denotes the predefined replay ratio. 

Aligned with the episodic fashion of meta-learning, we calculate the replay frequency
\begin{equation}
    R_F = \left\lceil \frac{R_I / s + 1}{b} \right\rceil
\end{equation}
Thus, every $R_F$-th episode can be considered as replay episode in a way that its query set does not consist of data from the stream, but from the episodic memory module $\mathcal{M}_{\mathcal{E}}$. 

\subsection{DRILL}

The DRILL architecture comprises four main elements, namely a dual-memory system of (1) an episodic memory module $\mathcal{M}_{\mathcal{E}}$ and (2) a semantic memory module $\mathcal{M}_{\mathcal{S}}$, and, following the original OML algorithm~\cite{Javed2019}, (3) a representation learning network (RLN) $h_{\phi}$ as well as (4) a prediction learning network (PLN) $g_{\boldsymbol{W}}$.

\begin{figure}[b!]
\includegraphics[width=\textwidth]{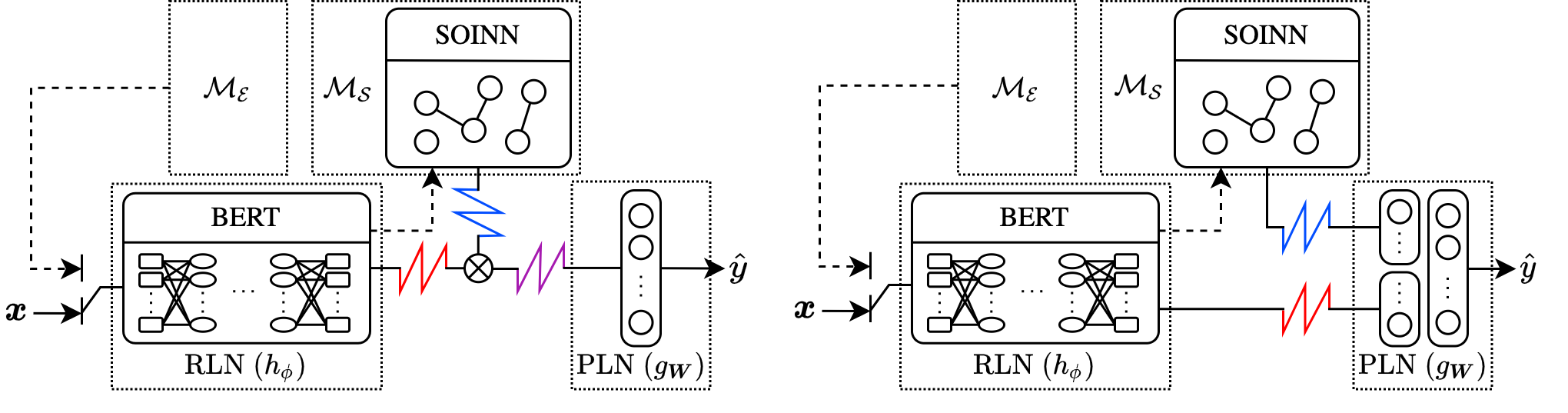}
\caption{Overview of the two variants $\text{DRILL}_{\text{M}}$ (left) and $\text{DRILL}_{\text{C}}$ (right). Latent representation signals retrieved from RLN are integrated with neural weight signals from $\mathcal{M}_{\mathcal{S}}$ either by multiplication ($\text{DRILL}_{\text{M}}$) or concatenation ($\text{DRILL}_{\text{C}}$). Input to the model is either an new observation $\boldsymbol{x}$ from the stream or an episodic replay sample from $\mathcal{M}_{\mathcal{E}}$.} \label{fig1}
\label{drill_overview}
\end{figure}

We use the SOINN+~\cite{Wiwatcharakoses2020} algorithm as semantic memory module $\mathcal{M}_{\mathcal{S}}$. Each neural unit is a $d$-dimensional real-valued vector. The network is parameterized by a pull factor $\eta$ denoting the influence of a new observation on neighboring nodes. For the sake of simplicity and as proposed by Wiwatcharakoses and Berrar~\cite{Wiwatcharakoses2020}, we set the pull factor to a constant value $\eta = 50$. 
Thus, our model $f_{\theta}$ optimizes for the set of parameters $\theta = \phi \, \cup \, \boldsymbol{W}$, consisting of parametrization $\phi$ from $\mathcal{X} \to \mathbb{R}^d$ of the RLN $h_{\phi}$ and $\boldsymbol{W}$ from $\mathbb{R}^d \to \mathcal{Y}$ of the PLN $g_{\boldsymbol{W}}$ respectively. 

Taking inspiration from the mammalian thalamus as a `gate to consciousness', we propose two DRILL variants that differ in how latent representations from the RLN are integrated with signals from the semantic memory $\mathcal{M}_{\mathcal{S}}$, translating its internal selective plasticity to the entire learning process.

The first variant, called \textit{Integration by Multiplication} ($\text{DRILL}_{\text{M}}$), can be described as follows: On receiving input $\boldsymbol{x}$, the model gates the activations $h_{\phi}(\boldsymbol{x})$ arriving from the RLN by multiplying them element-wise with a set of neural weights $\boldsymbol{w}_{\mathcal{S}}$ drawn from $\mathcal{M}_{\mathcal{S}}$ in a procedure described in subsection \ref{selfsupervisedsampling}. We express the model variant $\text{DRILL}_{\text{M}}$ as
\begin{equation}
    f^M_{\theta}(\boldsymbol{x}) = g_{\boldsymbol{W}} (\boldsymbol{w}_{\mathcal{S}} \cdot h_{\phi}(\boldsymbol{x}) )
\end{equation}
For the second variant called \textit{Integration by Concatenation} ($\text{DRILL}_{\text{C}}$), each of the $d$-dimensional signals $\boldsymbol{w}_{\mathcal{S}}$ and $h_{\phi}(\boldsymbol{x})$ retrieved from $\mathcal{M}_{\mathcal{S}}$ and the RLN respectively are reduced to half of their dimension $\frac{d}{2}$ and subsequently concatenated in a $d$-dimensional linear layer that is allocated to the PLN, as shown in Figure~\ref{drill_overview}. Thus, we derive the following model
\begin{equation}
    f^C_{\theta}(\boldsymbol{x}) = g_{\boldsymbol{W}} ( \, \left[ \boldsymbol{w}_{\mathcal{S}}, h_{\phi}(\boldsymbol{x}) \right] \, )
\end{equation}
where $\left[ \cdot , \cdot \right]$ denotes the concatenation operator. The meta-learning procedure for both DRILL variants works as follows: During inner-loop optimization of the $i$-th episode, the RLN is kept frozen while the PLN is fine-tuned using SGD with an inner-loop learning rate $\alpha$, such that
\begin{equation}
    \boldsymbol{W}' \leftarrow \text{SGD}(\mathcal{L}_i(\phi, \boldsymbol{W}), \mathcal{S}_i, \alpha)
\end{equation}
Subsequently, both RLN and PLN are fine-tuned on the query set $\mathcal{Q}_i$ during outer-loop optimization, such that all model parameters are updated using the Adam optimizer~\cite{kingma2015adam} with an outer-loop learning rate $\beta$ to give
\begin{equation}
    \theta' \leftarrow \text{Adam}(\mathcal{L}_i(\phi, \boldsymbol{W}'), \mathcal{Q}_i, \beta)
\end{equation}
For the RLN, we use the state-of-the-art transformer-based language model $\text{BERT}_{\text{BASE}}$~\cite{Devlin2019} with 12 transformer layers and $d = 768$ hidden dimensions. With $\text{DRILL}_{\text{M}}$, the PLN is a single linear layer with softmax activation that outputs the class probabilities, while a linear concatenation layer additionally precedes this layer with $\text{DRILL}_{\text{C}}$.  

\subsection{Self-Supervised Sampling}
\label{selfsupervisedsampling}

In contrast to the episodic memory $\mathcal{M}_{\mathcal{E}}$ that we solely use for experience replay, we use the semantic memory $\mathcal{M}_{\mathcal{S}}$ for generating high-quality representations, which influence the fine-tuning of the PLN. For every input sample $(\boldsymbol{x}_i, y_i)$, we initiate a competitive voting mechanism among all nodes in $\mathcal{M}_{\mathcal{S}}$ to determine the two neurons with neural weights $\boldsymbol{w}_{\mathcal{S}}^1$ and $\boldsymbol{w}_{\mathcal{S}}^2$ that have most frequently been best-matching units (BMUs) for class $y_i$. According to the original SOINN+ algorithm~\cite{Wiwatcharakoses2020}, the network node that lies closest to the input in Euclidean space is denoted as BMU. 

The two winners are then either multiplied element-wise ($\text{DRILL}_{\text{M}}$) or concatenated ($\text{DRILL}_{\text{C}}$) with the activations of the latent representation $h_{\phi}(\boldsymbol{x}_i)$ coming from the RLN, thus generating two inputs to the PLN from one output of the RLN. During the evaluation phase, only one signal from the winning node $\boldsymbol{w}_{\mathcal{S}}$ is retrieved from $\mathcal{M}_{\mathcal{S}}$ for the purpose of unambiguous label prediction by the PLN.

\section{Experiments}

\subsection{Benchmark Datasets}

We train our model sequentially on five text classification datasets by Zhang et al.~\cite{zhang2015} covering four different tasks: Sentiment analysis, news topic detection, question-and-answer classification, and ontology categorization. We summarize them in Table~\ref{dataset_table}. Following d'Autume et al.~\cite{dautume2019mbpa++}, the datasets are arranged in four randomized permutations reflecting the significant impact of task ordering on evaluation results.

\begin{table} 
\caption{The five balanced text classification datasets as in Zhang et al.~\cite{zhang2015}, each containing 7,600 test samples randomly drawn from the original datasets. The number of training samples differs depending on order position and imbalanced sampling strategy.}
\centering 
\begin{tabular*}{0.9\textwidth}{P{0.22\textwidth} P{0.18\textwidth} P{0.18\textwidth} P{0.06\textwidth} P{0.06\textwidth} P{0.06\textwidth} P{0.06\textwidth}}
\toprule 
 &  &  & \multicolumn{4}{c}{\textbf{Order Position}} \\
\cmidrule{4-7}
\multirow{-2}{*}{\textbf{\begin{tabular}[c]{@{}c@{}} Classification\\Domain\end{tabular}}} & \multirow{-2}{*}{\centering\textbf{Dataset}} & \multirow{-2}{*}{\textbf{Classes}} & \textbf{I} & \textbf{II} & \textbf{III} & \textbf{IV} \\
\midrule[0.3pt]
 & Amazon & & 4 & 4 & 3 & 3 \\
\multirow{-2}{*}{Sentiment} & Yelp & \multirow{-2}{*}{\begin{tabular}[c]{@{}c@{}}5 \\ (merged)\end{tabular}}  & 1 & 5 & 1 & 2 \\
\cmidrule[0pt]{1-3}
News Topic & AGNews & 4 & 2 & 3 & 5 & 1 \\
\cmidrule[0pt]{1-3}
Question Topic & Yahoo & 10 & 5 & 2 & 2 & 4 \\
\cmidrule[0pt]{1-3}
Ontology & DBPedia & 14 & 3 & 1 & 4 & 5 \\ 
\midrule[0.3pt] 
\multicolumn{2}{r}{\textbf{Total:}} & \textbf{33} & \multicolumn{4}{l}{} \\
\bottomrule 
\end{tabular*}
\label{dataset_table}
\end{table}

For evaluation, we follow prior work~\cite{holla2020,dautume2019mbpa++,sun2019lamol,sun2020distill} and randomly draw 7,600 samples from each of the five datasets, yielding a total test size of 38,000. However, we depart from the perfectly balanced and thus poorly realistic scenario of 115,000 training samples per dataset and instead apply progressive imbalancing as described in subsection~\ref{progressiveimbalancing} with $n_0
^{R}= 115,000$ and $n_0^{E} = 7,187$, thus providing a total training size of $222,812$ for either sampling strategy. 


\subsection{Baselines}

For performance evaluation, we compare our two proposed model variations $\textbf{DRILL}_{\textbf{M}}$ and $\textbf{DRILL}_{\textbf{C}}$ with the two performance leaders given a realistic single-epoch set-up without prior task-specific knowledge, i.e. \textbf{ANML-ER} and \textbf{OML-ER}~\cite{holla2020}. Just like our method, they use a pretrained $\text{BERT}_{\text{BASE}}$ language encoder. We further implement the lower bound for CL model performance, \textbf{SEQ}, in which we fine-tune both RLN and PLN on all tasks sequentially without any rehearsal. We also compare our methods with \textbf{REPLAY}, an extension of SEQ towards experience rehearsal with samples stored in an episodic memory. Finally, we train RLN and PLN jointly in a multitask set-up \textbf{MTL}, which we consider as an upper bound for CL model performance. For a fair comparison, we choose the same memory-write and rehearsal policies for REPLAY, ANML-ER, and OML-ER, as well as our two proposed DRILL variants. 

\subsection{Implementation Details}

Our experimental set-up consists of three independent runs on seeds 42-44, each run performed on the four order permutations and two sampling strategies respectively. Accordingly, the comparison results are averaged over all three runs. 

Due to computational limitations, we train all baseline models on normalized batches of size $s=8$ following the procedure of Ioffe and Szegedy~\cite{ioffe2015batchnorm} and optimize based on the cross-entropy loss on all 33 classes. We truncate the $\text{BERT}_{\text{BASE}}$ input sequences to length $448$ and set the buffer size $b = 6$. The inner-loop and outer-loop learning rates of the four meta-learning-based models $\text{DRILL}_{\text{M}}$, $\text{DRILL}_{\text{C}}$, OML-ER, and ANML-ER are set to $\alpha = 8\mathrm{e}{-3}$ and $\beta = 1.5\mathrm{e}{-5}$ respectively. The learning rate of all remaining baselines SEQ, REPLAY, and MTL is set to $1\mathrm{e}{-5}$. 

All models are trained for a single epoch, whereas MTL is trained for two epochs. The probability of storing an observation in the episodic memory module $\mathcal{M}_{\mathcal{E}}$ is governed by the maximum write probability $p_{\mathcal{E}} = 0.8$. The $p_{\mathcal{E}}$ is inversely proportional to the expansion or reduction for all rehearsal-based models, restoring class balance within $\mathcal{M}_{\mathcal{E}}$. The learning rates and $p_{\mathcal{E}}$ are derived using a Parzen–Rosenblatt estimator~\footnote{CometML Hyperparameter Optimizer: \url{https://www.comet.ml/}}. The hyperparameter optimization is applied to OML-ER as the representative model for all meta-learning-based approaches and SEQ for inferring the learning rate of the remaining models. Both OML and SEQ are trained on the full dataset (without expansion or reduction) with order I and random seed 42. With both DRILL architectures, the unsupervised SOINN+ algorithm is performed as described in the original paper~\cite{Wiwatcharakoses2020}, including setting the pull factor $\eta = 50$. 

We follow the rehearsal and evaluation strategies adopted by Holla et al.~\cite{holla2020}, setting $R_I = 9,600$ and $r = 1\%$, such that we draw 96 samples from $\mathcal{M}_\mathcal{E}$ after observing 9,600 samples from the data stream. The evaluation of the four meta-learning models is performed by generating five episodes, each containing the test datasets as query sets. All baseline models were trained on an NVIDIA TITAN RTX with 24GB VRAM and 64GB RAM. The training time ranges between 1 and 7 hours, depending on the model and number of observations.

\section{Results}

\subsection{Imbalanced Lifelong Text Classification}

\begin{table} 
\caption{Text classification $F_1$ scores on four permutations of task orders and progressive expansion ($E$) and progressive reduction ($R$) sampling respectively. The two rightmost columns denote the macro-average and standard deviation across all orderings and sampling strategies.}
\centering 
\begin{tabular*}{0.938\textwidth}{p{0.15\textwidth} P{0.07\textwidth} P{0.07\textwidth} P{0.07\textwidth} P{0.07\textwidth} P{0.07\textwidth} P{0.07\textwidth} P{0.07\textwidth} P{0.07\textwidth} P{0.07\textwidth} P{0.07\textwidth}}
\toprule 
& \multicolumn{4}{c}{\textbf{Order ($E$)}} & \multicolumn{4}{c}{\textbf{Order ($R$)}} \\ 
\cmidrule(rl){2-5} 
\cmidrule(rl){6-9}
\textbf{Method} & \textbf{I} & \textbf{II} & \textbf{III} & \textbf{IV} & \textbf{I} & \textbf{II} & \textbf{III} & \textbf{IV}  &$\mathbf{\mu}$ &$\mathbf{\sigma}$\\ 
\midrule[0.3pt] 

$ \text{SEQ} $ &  $17.4$  &  $27.6$  &  $26.6$  &  $21.0$  &  $23.7$  &  $32.7$  &  $28.8$  &  $25.0$   &  $25.4$ & $4.9$ \\
$ \text{REPLAY} $ &  $55.3$  &  $67.9$  &  $58.6$  &  $\mathbf{65.7}$  &  $44.2$  &  $57.7$  &  $53.5$  &  $37.0$  &  $55.0$ & $10.5$  \\
$ \text{ANML-ER} $ &  $66.7$  &  $\mathbf{70.5}$  &  $55.0$  &  $62.9$  &  $57.0$  &  $58.6$  &  $62.7$  &  $45.2$  &  $59.8$  & $8.8$ \\
$ \text{OML-ER} $ &  $\mathbf{70.2}$  &  $64.9$  &  $52.2$  &  $64.4$  &  $56.0$  &  $\mathbf{62.0}$  &  $\mathbf{66.5}$  &  $48.7$ & $60.6$  & $8.1$  \\
$ \text{DRILL}_{\text{M}} $ &  $23.2$  &  $36.5$  &  $37.1$  &  $37.6$  &  $58.0$  &  $51.7$  &  $41.0$  &  $\mathbf{50.4}$  & $41.9$ &  $ 13.7$  \\
$ \text{DRILL}_{\text{C}} $ &  $68.4$  &  $68.1$  &  $\mathbf{59.1}$  &  $65.5$  &  $\mathbf{61.8}$  &  $61.6$  &  $62.9$  &  $49.5$ &  $\mathbf{62.1}$  & $6.2$  \\
\midrule[0.3pt] 
\midrule[0.3pt] 
$ \text{MTL} $ &  $77.9$  &  $78.7$  &  $76.2$  &  $76.7$  &  $77.7$  &  $76.4$  &  $78.3$  &  $78.2$  &  $77.5$ & $1.0$ \\



\bottomrule 
\end{tabular*}
\label{result_table}
\end{table}

Unlike prior work, we report $F_1$ scores rather than macro-averaged classification accuracy due to the unbalanced nature of the training data. Our main results are summarized in Table~\ref{result_table}.

\begin{figure}
\includegraphics[width=\textwidth]{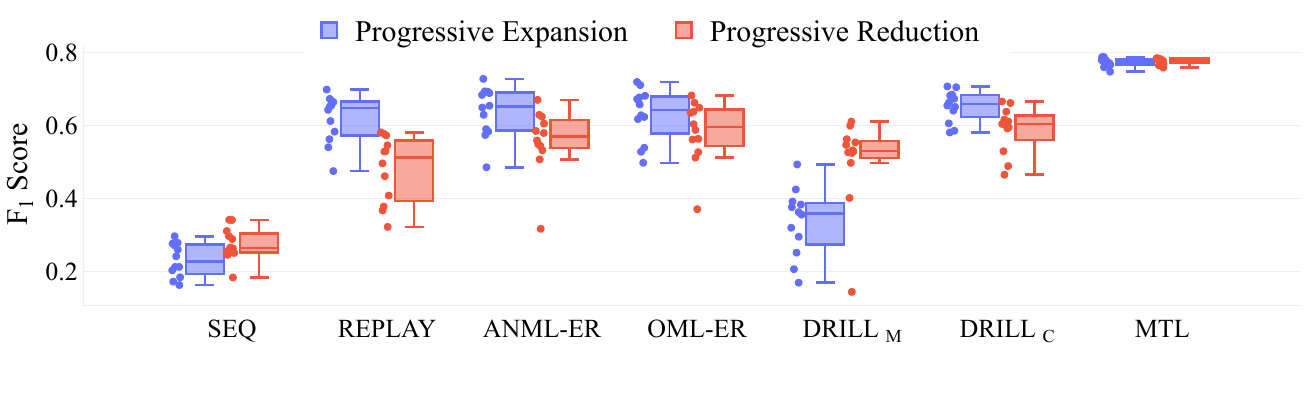}
\caption{$F_1$ scores of all comparison models aggregated across three seeds and four orderings. Sequential (SEQ) and multitask (MTL) learning can be viewed as lower and upper bound for model performance respectively.} 
\label{class_results}
\end{figure}

Our $\text{DRILL}_{\text{C}}$ variant outperforms existing methods in terms of higher overall average performance and higher median under equal conditions (the latter is depicted in Figure~\ref{class_results}). In addition, it has a significantly smaller variance than all other replay-based comparison methods, thus demonstrating its robustness to the order of training data and the imbalancing strategy. Consequently, it narrows the gap to the upper bound of multitask learning. 

Interestingly, the $\text{DRILL}_{\text{M}}$ method is trailing the current models with respect to absolute performance. Yet, it provides a smaller variance for progressively expanded data than all other baselines except SEQ, exhibiting robustness against undersampled classes at the beginning of training. The enormous performance difference of our two DRILL variants motivates a more detailed analysis of the impact of knowledge integration mechanisms from RLN and $\mathcal{M}_{\mathcal{S}}$.

\subsection{Knowledge Integration Mechanisms}

Although the introduction of class-representative signals drawn from semantic memory yields greater robustness under a realistic training scenario, the overall model performance varies greatly depending on how the latent signals retrieved are integrated during training. The relatively poor performance of $\text{DRILL}_{\text{M}}$ could be attributed to the multiplicative gating mechanism that we adopted from the original ANML algorithm~\cite{Beaulieu2020}. The ANML is designed so that `gating parameters' of preceding layers are learned in a supervised fashion, which is in contrast to the unsupervised nature of SOINN+.

Conversely, with $\text{DRILL}_{\text{C}}$, signals from RLN are enriched with those from the SOINN rather than fused, allowing for better linear separation, thus resulting in an increase of model performance. From this, we conclude that the concatenation of modalities in our training scenario provides a better knowledge retention strategy.

\subsection{Self-Organized Networks in NLP}

A generally known problem of self-organizing networks is that they capture the entire evolution of hidden representations in feature space along with obsolete knowledge and are therefore unsuitable for training on shifting latent distributions. With the DRILL architecture, we overcome this problem by freezing the RLN parameters during inner-loop optimization and by the choice of our retrieval strategy for neural weight signals coming from the SOINN. The former leads to a more stable latent data distribution over a longer period. The latter ensures that neural units residing in the current input distribution are more likely to be considered as high-quality class representatives.

As this is the first work to combine a self-organizing neural architecture with a transformer-based language model in a CL setting, we advocate further exploring such set-ups in future work. This is due to the intrinsic ability of the SOINN and its various extensions to be applicable in an infinite learning setting with an unlimited number of tasks. The model can additionally handle partially annotated data, setting the basis for semi-supervised LLL scenarios.

\section{Conclusion and Future Work}

In this work, we introduce a novel, more challenging continual learning set-up with imbalanced data. We further propose Dynamic Representations for Imbalanced Lifelong Learning (DRILL), a neuroanatomically inspired CL method which combines a state-of-the-art language model with a self-organizing neural architecture. It outperforms current baselines, yet is more stable against data ordering and imbalancing. Thus, the fusion of supervised language models with unsupervised clustering algorithms has proven effective for lifelong learning methods, further narrowing the gap to multitask learning approaches. DRILL achieves the best results on imbalanced data, with the least overall variance in comparison to other meta-learning-based lifelong learning approaches. For future work, we plan to extend our model towards infinite learning of an unknown number of tasks as well as sequence-to-sequence learning.

\section*{Acknowledgements}
We would like to thank Dr. Cornelius Weber (University of Hamburg) and Katja K\"{o}sters (University of Hamburg) for their feedback and suggestions.
The authors gratefully acknowledge partial support from the German Research Foundation (DFG) under Project CML (TRR-169).

%
%
%
\bibliographystyle{splncs04}
\bibliography{DRILL_Paper}
\end{document}